\documentclass[accepted]{uai2024} 
                        
\usepackage[american]{babel}

\usepackage{natbib} 
\usepackage{mathtools} 
\usepackage{booktabs} 
\usepackage{tikz} 
\usepackage{amsmath,amssymb,amsfonts}
\usepackage{algorithm}
\usepackage{algpseudocode}
\usepackage{tabularx,colortbl}
\usepackage{caption,subcaption}
\usepackage{enumitem}
\usepackage{verbatim}
\begin{document}
\title{Echoes of Socratic Doubt: Embracing Uncertainty in\\Calibrated Evidential Reinforcement Learning}

%
%
\author[1]{\href{mailto:<astutt2@uic.edu>?Subject=Your Reinforcement Learning Paper}{Alex~Christopher~Stutts}{}}
\author[1]{Danilo~Erricolo}
\author[2]{Theja~Tulabandhula}
\author[1]{Amit~Ranjan~Trivedi}
\affil[1]{%
    Department of Electrical and Computer Engineering\\
    University of Illinois Chicago\\
    Chicago, Illinois, USA
}
\affil[2]{%
    Department of Information and Decision Sciences\\
    University of Illinois Chicago\\
    Chicago, Illinois, USA
}

\maketitle

\begin{abstract}
  We present a novel statistical approach to incorporating uncertainty awareness in model-free distributional reinforcement learning involving quantile regression-based deep Q networks. The proposed algorithm, \emph{Calibrated Evidential Quantile Regression in Deep Q Networks (CEQR-DQN)}, aims to address key challenges associated with separately estimating aleatoric and epistemic uncertainty in stochastic environments. It combines deep evidential learning with quantile calibration based on principles of conformal inference to provide explicit, sample-free computations of \emph{global} uncertainty as opposed to \emph{local} estimates based on simple variance. This approach overcomes limitations of traditional methods in computational and statistical efficiency and handling of out-of-distribution (OOD) observations. Tested on a suite of miniaturized Atari games (i.e., MinAtar), CEQR-DQN is shown to surpass similar existing frameworks in scores and learning speed. Its ability to rigorously evaluate uncertainty improves exploration strategies and can serve as a blueprint for other algorithms requiring uncertainty awareness.
\end{abstract}

\section{Introduction}\label{sec:intro}
\begin{figure}[!t]
  \centering
  \includegraphics[width=\linewidth]{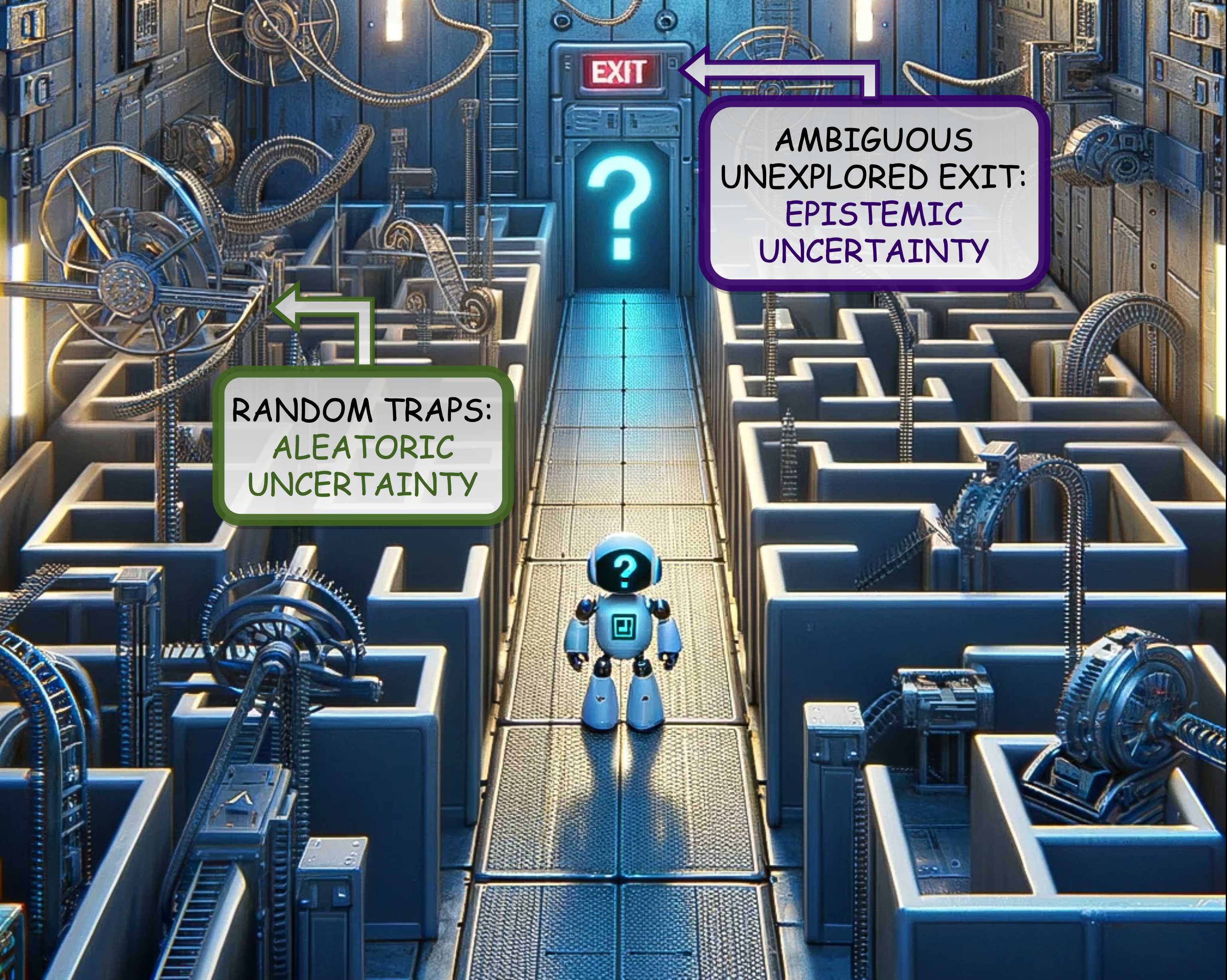}
  \caption{\textbf{Deep Uncertainty-Aware Reinforcement Learning:} In this study, we present a novel solution for separately quantifying aleatoric and epistemic uncertainty in distributional reinforcement learning involving deep Q networks. The proposed framework combines deep evidential learning with calibrated quantile regression based on conformal inference to significantly enhance agent exploration through uncertainty-aware action selection. The graphic above depicts a cautious, uncertainty-aware robot agent trying to escape a dangerous maze laden with traps and ambiguous paths; it was generated with the assistance of AI and post-edited.}
  \label{fig:overview}
\end{figure}
Artificial intelligence (AI) has seen incredible growth in recent years in terms of interest, research, applications, and performance. Anyone with even basic knowledge of the technology has been able to utilize its tools for processing unprecedented amounts of data from audio, visual, and/or textual sources. However, the insights from the models are only as good as the data they consume regardless of the level of supervision in training. Data, especially from the real world, naturally and often randomly contain noise and inconsistencies, which lead to uncertainty in model predictions and can potentially degrade performance. The two main types of uncertainty that affect the models are referred to as \emph{aleatoric} and \emph{epistemic}, where the former is the result of random distortions in the data and is typically unresolvable, and the latter is the result of insufficient knowledge, which can be mitigated with more data and training.

Deep reinforcement learning (DRL) is a subset of machine learning (ML) under the umbrella of AI that involves using neural networks to solve sequential decision-making problems that require interaction between an agent and an environment. Recent applications of DRL include drone racing \citep{droneracing2023}, healthcare diagnosis and treatment \citep{healthcareRL2021}, and chip design \citep{chipdesignRL2021}, which are all distinct tasks. As such, there are several types of DRL algorithms, although in this study we focus on deep Q networks with discrete action spaces \citep{Mnih2015}. In regards to processing data, a key difference between most DRL algorithms and other ML algorithms is their foundation in the Markov Decision Process (MDP) \citep{Puterman1994}, a mathematical framework for discrete-time decision-making when the desired outcome (e.g., maximum cumulative reward) is stochastic yet partially amenable to a user. Accordingly, DRL is susceptible to uncertainty, which is reflected directly in each interaction and the end result. However, due to the nature of the algorithm's interactive data processing, one is given explicit parametric control over whether to exploit or explore the environment when seeking rewards, which can be manipulated with uncertainty awareness. Unlike most supervised ML tasks, where uncertainty awareness has seen the most interest lately, reinforcement learning can directly and explicitly incorporate reliable uncertainty measures as feedback in the exploitation/exploration trade-off and choosing actions.

There has been a number of studies exploring solutions to important challenges in uncertainty-aware reinforcement learning \citep{lockwood2022}. One challenge involves ensuring statistical rigor and efficiency in estimating both aleatoric and epistemic uncertainty, whereas most previous works have focused only on epistemic uncertainty due to the intractability of conventional uncertainty estimation methods in DRL (e.g. Bayesian sampling). Another challenge involves accurately separating the two quantities since epistemic uncertainty is often a function of aleatoric uncertainty; an agent unaware of sources of aleatoric uncertainty in the environment (e.g., random traps) may take actions that neglect exploring related sources of epistemic uncertainty (e.g., shortcuts), hindering its ability to gain knowledge of potentially better rewards. Sources of either uncertainty can generally yield positive or negative rewards, but the only way the agent will learn this effectively is through uncertainty awareness, the very essence of socratic doubt. An example of this relationship is given in Fig.~\ref{fig:overview}. A third challenge involves properly handling observations that significantly deviate from the norm (i.e., out-of-distribution or OOD) \citep{yang2021oodsurvey}, which tend to negatively affect both algorithm performance and traditional uncertainty measures that rely on restrictive assumptions such as Gaussianity. Lastly, unlike traditional ML, the \emph{target} in DRL is not fixed (whether it is a single value or a distribution), and samples are not considered traditionally independent and identically distributed (i.i.d) so certain theoretical bases become invalid. Therefore, the uncertainty quantification in DRL needs to be highly adaptive and flexible, yet stable and capable of empirically accommodating violations of typical critical assumptions.

To address these challenges of uncertainty estimation in DRL, we propose a novel combination of properties of conformal inference \citep{vovk2005,shafer2007,angelopoulos2021} and deep evidential learning \citep{sensoy2018,amini2020} to further enhance the usage of quantile regression as a basis for distributional reinforcement learning involving deep Q networks (i.e., QR-DQN) \citep{bellemare2017distributional,dabney2017distributional}. Conformal inference is a simple, widely applicable, and scalable statistical framework for constructing reliable prediction intervals using a finite set of data with guaranteed marginal coverage of the true realization of interest, subject to a predetermined confidence rate and a relevant calibration dataset. It excels at adaptation to complex, heteroskedastic data and fortifies estimation of aleatoric and epistemic uncertainty when paired with an inherently uncertainty-aware algorithm such as quantile regression \citep{romano2019}, without making strong theoretical assumptions. However, conformal inference alone can be overly conservative and does not perform well on new OOD observations without additional training and/or calibration. On the other hand, evidential learning involves modeling a higher-order probability distribution placed over the likelihood function through learned parameters, which bypasses the need for repetitive sampling typically attributed to Bayesian techniques and allows one to explicitly compute aleatoric and epistemic uncertainty separately. In practice, it is equally suitable as an accurate uncertainty estimation method, but is much superior in handling OOD observations. Nonetheless, evidential learning is more computationally complex and therefore requires careful optimizations to maintain stability depending on the data. By combining the two complementary concepts into one algorithm, we can leverage their benefits and mitigate their weaknesses.

Thus, the proposed study makes the following key contributions:
\begin{itemize}[itemsep=2pt,topsep=2pt,leftmargin=10pt]
\item We introduce a novel framework that combines elements of conformal inference and deep evidential learning to improve the usage of quantile regression in deep distributional reinforcement learning. The algorithm incorporates separate, explicit, and statistically robust computations of aleatoric and epistemic uncertainty as feedback for balancing exploration and exploitation in action selection.
\item In utilizing properties of conformal inference, the estimated quantiles of the distribution of returns over each action taken in the environment are \emph{calibrated} with the distributional Bellman target \citep{bellemare2017distributional,dabney2017distributional} according to a desired marginal coverage of $90\%$. This enhances the accuracy of the Q-value quantiles as training samples are selected from experience replay. Additionally, the estimated quantiles of each learned evidential distribution (which are used to compute uncertainty) are similarly calibrated, allowing the algorithm to tighten action selection and maintain high score performance well into training.
\item Combining calibrated quantile regression with deep evidential learning enhances the algorithm's uncertainty awareness by providing \emph{global} network uncertainty estimates over the entire known sample space as opposed to \emph{local} estimates based solely on the variance and expectation of Q-value quantiles, which can be biased. As a result, the proposed CEQR-DQN algorithm achieves meets or exceeds scores on benchmarks such as MinAtar \citep{young19minatar} set by similar previous work with notably faster learning.
\end{itemize}

In Section~\ref{sec:related}, we briefly review work related to our proposed solution for uncertainty awareness in DRL. In Section~\ref{sec:main}, we discuss the algorithm and underlying theory in detail. In Section~\ref{sec:results}, we present simulation results. Lastly, Section~\ref{sec:conclusion} concludes the study.

\section{Related Work}\label{sec:related}
The study of uncertainty awareness in deep reinforcement learning has remained under active investigation for several years, in part due to its undeniable impact on returns. Different approaches to the topic have attempted to address the many challenges associated with parameterizing, stabilizing, and separating uncertainty estimates subject to the stochastic nature of reinforcement learning. In this study, we build upon a particular thread of influential uncertainty-aware work that involve deep Q learning, a simple, model-free DRL algorithm that focuses on approximating the Q-value function given a discretized action space (i.e., DQN) \citep{Mnih2015}.

Bootstrapped DQN \citet{osband2016} was one of the first algorithms to address exploration efficiency through incorporating a sense of uncertainty in choosing actions. As an alternative to Thompson sampling \citep{thompson1933}, their approach involves using multiple neural network heads with random initializations to provide a notion of uncertainty and encourage an agent to dynamically choose actions based on knowledge (or lack thereof) of potential rewards, as opposed to an $\epsilon$-greedy policy. However, redundant approximations require computational overhead and this method of uncertainty estimation is rather rudimentary as random initilizations do not guarantee diversity or generalizability.

Double Uncertain Value Network (DUVN) \citet{moerland2017}, the first distributional DRL algorithm to separate and propagate both aleatoric and epistemic uncertainty, offers a more nuanced exploration strategy in comparison to bootstrapped DQN, which leads to better overall outcomes. Nonetheless, it is similarly unsophisticated in its uncertainty estimation as it uses dropout as an indicator of epistemic uncertainty and assumes that the return distribution (from which aleatoric uncertainty is extracted) is Gaussian. Furthermore, it is also computationally expensive as it requires careful optimization of two separate networks for the estimation, neglecting to account for their interaction.

Uncertainty-Aware DQN (UA-DQN) \citet{clements2020} builds upon the previous work by adopting the quantile regression-based distributional DRL solution proposed in \citep{dabney2017distributional}, establishing two bootstrapped auxiliary networks to assess the uncertainty quantities, and intelligently integrating MAP sampling to enhance uncertainty awareness. It is shown to properly estimate the uncertainties and be superior in performance, but is subject to the same vulnerabilities as other methods that rely on redundancy and random sampling.

Acknowledging the advantages and disadvantages of the aforementioned algorithms, the proposed work is inspired to maintain proper disentanglement in estimating \emph{global} aleatoric and epistemic uncertainty without bootstrapping auxiliary networks. Additionally, we seek to provide a more statistically rigorous algorithm that can empirically handle OOD observations even in highly complex environments.

\section{Uncertainty-Aware Reinforcement Learning}\label{sec:main}
This section details the proposed framework for deep uncertainty-aware distributional reinforcement learning involving calibrated quantile regression and evidential learning (i.e., calibrated evidential quantile regression in deep Q networks or CEQR-DQN). The scalable architecture is shown in Fig.~\ref{architecture} in the appendix; it adapts the original DQN feature extractor~\citep{Mnih2015}, and includes separate linear network heads for outputting quantiles for actions and the parameters of the empirically learned evidential distributions. We designed this model and the underlying methodology to be closely comparable to \citep{clements2020} and easy to adapt to other reinforcement learning solutions.

In accordance with the original QR-DQN \citep{dabney2017distributional}, our algorithm seeks to learn the distribution of returns $Z^{\pi}(x,a)$ (i.e., the state-action value distribution) upon taking action $a$ in state $x$ according to a policy $\pi$ using quantile regression. It follows the Markov Decision Process of modeling interactions between the agent and the environment based on the variables $X$, $A$, $R$, $P$, and $\gamma$, where $X$ represents the state space, $A$ represents the action space, $R$ represents the stochastic reward function, $P$ represents the probability of transitioning states $x\rightarrow x'$ after taking an action $a$, and $\gamma$ is the discount factor controlling the weight of future $Z^{\pi}(x,a)$ in loss calculations. Regarding loss, the value distribution is approximated via the temporal difference-based (TD) distributional Bellman target \mbox{$T^{\pi}Z(x,a)\stackrel{d}{:=}R(x,a)+\gamma Z(x',a')$} where $a'\sim$ argmax$_{a'}\mathbb{E}[Z(x',a')]$ and $x'\sim P(\cdot\,|\,x,a)$. It is parameterized with a quantile distribution that is learned by minimizing the quantile Huber loss 
\begin{equation}
\mathcal{\rho}^{\kappa}_{q} = |q-\mathbb{I}\{e<0\}|\frac{\mathcal{L}_{\kappa}(e)}{\kappa}
\end{equation}
with real-valued quantile functions of the cumulative distribution function $F$ (i.e., $\theta=F^{-1}_{Z}(\hat{q})$), where
\begin{equation}
\mathcal{L}_{\kappa}(e)=\mathbb{I}\{|e|\leq\kappa\}\frac{1}{2}e^2+\mathbb{I}\{|e|>\kappa\}\kappa\left(|e|-\frac{1}{2}\kappa\right)
\end{equation}
is the Huber loss \citep{huber1964}, $q\in[0,1]$ represents a target quantile, $\hat{q}$ represents an estimated quantile, $e$ is error, $\mathbb{I}$ is the indicator function, and $\kappa$ is a hyperparameter used to assert squared error within $[-\kappa,\kappa]$ distance of zero, or standard error otherwise. Quantile huber loss is a smoother version of the traditional quantile regression loss that is convex and asymmetric, penalizing overestimation and underestimation of an estimated conditional quantile.

Applying this framework to deep Q learning, \mbox{$e\equiv TZ_{\theta}(x,a)-\theta_{i}(x,a)$} where $Z_{\theta}$ represents the parameterized quantile distribution of returns mapping state-action pairs to a uniform probability distribution of $N$ quantiles with uniform weight and cumulative probabilities $\tau_{i}=\frac{1}{N}$, $i=1,\dots,N$ given $\theta: X\times A\rightarrow\mathbb{R}^N$. Accordingly, the network outputs $N$ quantiles for each discretized action and the resultant quantile regression loss is calculated as 
\begin{equation}
\mathcal{L}_{qr}=\sum_{i=1}^{N}\mathcal{\rho}^{\kappa}_{\hat{q_{i}}}(TZ_{\theta}(x,a)-\theta_{i}(x,a))\text{.}
\end{equation}
Notably, we approximate only one target value distribution $Z_{\theta}$ per update as opposed to multiples contained in a set of all possible distributions $Z_{Q}$ as is done in ensemble methods.

Different methods for DRL, such as policy gradient, actor-critic, and model-based methods, can incorporate uncertainty measures in various other aspects of agent-environment interaction. QR-DQN is an off-policy method and, as such, must incorporate uncertainty measures related to the value distribution. Therefore, we focus on uncertainty estimation in action selection as it is the primary component in maximizing returns. As opposed to other works following a similar procedure, the proposed method utilizes evidential learning to evaluate and exploit uncertainty in choosing actions that maximize the target value distribution. Actions are selected via Thompson sampling of a multivariate normal distribution based on the expectation of action quantiles and the covariance between the evidential uncertainty associated with each action.

For quick reference, the pseudocode for action selection, which includes evidential learning, is given in Algorithm~\ref{alg:cap} and explained in Section~\ref{sec:compiled_theory}.
\begin{algorithm}[!t]
\caption{Calibrated Evidential QR-DQN Action Choice}\label{alg:cap}
\begin{algorithmic}
\Require $N$ quantiles, action space $A$, state space $X$, network $\mathbb{N}$, evidential distribution parameters \mbox{$G=\{\gamma,v,\alpha,\beta\}_{evi}$}, and uncertainty hyperparameters $\lambda_{al}$ and $\lambda_{ep}$.
\Ensure $\{\lambda_{al}, \lambda_{ep},v,\beta\}>0$; \{$N, \alpha\}>1$
\State Gather quantiles of actions and evidential parameters given state $x$ in $X$ => $A_{N},G_{N}=\mathbb{N}(x)$
\For{$a$ in $A$}
\State Calculate action mean over N: $\mu=\frac{1}{N}\sum_{i=1}^N a_{i}$
\State Calculate evidential SD: $\sigma_{evi}=\sqrt{\left(\frac{\beta}{v(\alpha-1)}\right)}$
\State Calculate uncertainty for evidential $5^{th}$ percentile:
\State $u^{5^{th}}_{evi}=[|(\gamma^{5^{th}}_{evi}+\sigma^{5^{th}}_{evi})-(\gamma^{5^{th}}_{evi}-\sigma^{5^{th}}_{evi})|]$
\State Calculate uncertainty for evidential  $95^{th}$ percentile:
\State $u^{95^{th}}_{evi}=[|(\gamma^{95^{th}}_{evi}+\sigma^{95^{th}}_{evi})-(\gamma^{95^{th}}_{evi}-\sigma^{95^{th}}_{evi})|]$
\State Calculate total epistemic uncertainty:
\State $\psi_{ep}=\mathbb{E}_{N}[\frac{1}{2}(u^{5^{th}}_{evi}+u^{95^{th}}_{evi})]$
\State Calculate total aleatoric uncertainty:
\State $\psi_{al}=\mathbb{E}_{N}[|\gamma^{95^{th}}_{evi}-\gamma^{5^{th}}_{evi}|]$
\EndFor
\State Risk aversion (optional): $\mathcal{M}\leftarrow\mathcal{M}-\lambda_{al}\Psi_{al}$
\State Draw Thompson sample: \mbox{$S_{A}\sim\mathcal{N}(\mathcal{M},\lambda_{ep}\Psi_{ep})$}
\State (where $\mathcal{M}=\{\mu_{1},\dots,\mu_{A}\}$; $\Psi_{el,ap}=\{\psi_{1},\dots,\psi_{A}\}$)
\State \hspace{-1em}\textbf{Output action:} argmax$_{a}[S_{A}]$
\end{algorithmic}
\end{algorithm}

\subsection{Deep Evidential Learning for Action Choice}\label{sec:DEL}
Deep evidential learning (DEL) is a relatively recent methodology for uncertainty estimation in deterministic neural networks, yet it already has an extensive history of development in various applications such as classification, regression, \emph{etc.} \citep{ulmer2023prior}. It poses learning as an evidence gathering problem \citep{sensoy2018,malinin2018predictive,amini2020} where each training sample increases knowledge of a higher-order evidential distribution whose priors are placed over the likelihood function as opposed to network weights, superseding the costly procedure of conventional Bayesian techniques in neural networks. This allows the network to parameterize the evidential distribution for direct computation of global epistemic and aleatoric uncertainty, making the process entirely tractable and sample-free. A key advantage of DEL is its interpretation and handling of out-of-distribution observations. OOD samples almost always cause significant degradation in deterministic predictions due to insufficient knowledge and a lack of a valid OOD training dataset. However, DEL is highly adaptive and resistant against such decline in performance, accurately inflating uncertainty upon certain observations and penalizing evidence upon severe estimation errors.

In this study, we adopt the deep evidential regression method proposed in \citep{amini2020} with modifications based on \citep{hüttel2023deep} for uncertainty estimation in DRL action selection. Deep evidential regression jointly learns a continuous target and both aleatoric and epistemic uncertainty by training a network to predict parameters of a higher-order evidential conjugate prior distribution placed over the parameters of an unknown Gaussian distribution from which targets belong. This method is extended with Bayesian quantile regression in \citep{hüttel2023deep} to guarantee proper modeling of non-Gaussian target distributions, effectively relaxing the typically critical assumption of Gaussianity in estimating aleatoric uncertainty. With these methods, our custom QR-DQN algorithm can fruitfully leverage the benefits of evidential learning with minimal concern over violations of restrictive assumptions and other statistical challenges associated with reinforcement learning and the stochastic MDP. Additionally, our efforts to calibrate the quantiles related to approximating $Z_{\theta}$ can similarly be applied to the quantiles integrated with the evidential parameters albeit for different objectives yielding distinct advantages, which is explained further in Section~\ref{sec:calibration}.

In order to infer the parameters of the evidential prior placed over the maximum likelihood function to maximize the likelihood of observing true values of the target distribution and thus assess aleatoric uncertainty as well as epistemic uncertainty, the regression procedure assumes a hierarchical Bayesian structure. The target distribution $\mathcal{N}$ is parameterized with ($\mu,\sigma^{2}$), but as is usually the case in deep Bayesian inference, $\mu$ and $\sigma^{2}$ are not directly determinable axiomatically, so their individual distributions must be estimated instead which involves placing a normal prior on $\mu$, and an Inverse-Gamma prior on $\sigma^{2}$: 
\begin{equation}
\mu\sim\mathcal{N}(\gamma,\sigma^{2},v^{-1});~\sigma^{2}\sim\Gamma^{-1}(\alpha,\beta)\text{,}
\end{equation}
where $\gamma\in\mathbb{R}$, $\{v,\beta\}>0$, $\alpha>1$, and $\Gamma^{-1}(\cdot)$ is the Inverse-Gamma distribution. This allows for tractable, sample-free Bayesian inference and modeling of non-Gaussian distributions. When united, these distributions formulate the Normal-Inverse-Gamma (NIG) evidential prior distribution (i.e., normal conjugate prior) which is shown in (5) and parameterized with learnable parameters $G=\{\gamma,v,\alpha,\beta\}$ where generally $\gamma$ represents the expected sample mean, $v$ represents the precision, $\alpha$ represents the shape, and $\beta$ represents the scale of the distribution \citep{bishop2006,amini2020,hüttel2023deep}:
\begin{equation}
\begin{split}
p(\mu,\sigma^{2}|\{\gamma,v,\alpha,\beta\})& =p(\mu|G)p(\sigma^{2}|G) \\
& =\frac{\beta^{\alpha}\sqrt{v}}{\Gamma(\alpha)\sqrt{2\pi\sigma^{2}}}\left(\frac{1}{\sigma^{2}}\right)^{\alpha+1} \\
& \times\text{exp}\left(\frac{-2\beta+v(\gamma-\mu)^{2}}{2\sigma^{2}}\right)\text{.}
\end{split}
\end{equation}

In addition to the interpretation of the evidential NIG parameters based on \emph{virtual observations} \citep{jordan2009} in support of learned evidence in \citep{amini2020}, \citet{hüttel2023deep} interprets $v$ and $\alpha$ as measures of evidential confidence on the normal prior of $\mu$ whereas $\beta$ indicates variance. Thus, total evidence (i.e., model confidence) is defined as $\Phi=2v+\alpha+\frac{1}{\beta}$ and is minimized in a regularizing loss function to penalize confidence and increase uncertainty when the network makes poor predictions. In the context of quantile regression, this evidence is scaled with \emph{tilted loss} in the loss function instead of absolute error in order to inflate confidence when a target quantile $q$ predicted for target $y$ is greater than $0.5$ and is overestimated versus underestimated (and \emph{vice versa} for $q<0.5$) via the implicit relationship $\beta\gg v,\alpha$:
\begin{equation}
\begin{split}
\mathcal{L}_{reg}=\Phi&[\mathbb{I}\{y\geq\hat{y}\}q (y-\hat{y}) \\
& +\mathbb{I}\{y<\hat{y}\}(1-q)(\hat{y}-y)]\text{,} \\
\end{split}
\end{equation}
where $y$ is an element of all true values $Y$, and $\hat{y}$ is the prediction.

Learning the evidential NIG distribution parameters also involves training a network to maximize evidence in support of observations. To maximize evidence, the network must maximize the marginal probability $p(y\,|\,G)$, which has a solution in the Student-t distribution St($\cdot$) evaluated at $y$ with location $\mu_{St}$, scale $\sigma^{2}_{St}$, and degrees of freedom $v_{St}$:
\begin{equation}
p(y\,|\,G)=\text{St}\left(y;\gamma,\frac{\beta(1+v)}{v\alpha},2\alpha\right)\text{.}
\end{equation}
This solution for maximizing model evidence is translated into another loss function as minimizing the negative log likelihood of the Student-t distribution \citep{amini2020,hüttel2023deep}:
\begin{equation}
\begin{split}
\mathcal{L}_{NLL}& =\frac{1}{2}\log\left(\frac{\pi}{v}\right)-\alpha\log(\Omega) \\
& +\left(\alpha+\frac{1}{2}\right)\log((y-\gamma)^{2}v+\Omega) \\
& +\log\left(\frac{\Gamma(\alpha)}{\Gamma\left(\alpha+\frac{1}{2}\right)}\right)\text{,}
\end{split}
\end{equation}
where $\Omega=2\beta(1+v)$. Thus, the complete loss function associated with evidential regression is
\begin{equation}
\mathcal{L}_{evi} = \mathcal{L}_{NLL}+\lambda_{reg}\mathcal{L}_{reg}
\end{equation}
where $\lambda_{reg}$ is a hyperparameter to balance inflation of uncertainty with learning. 

Finally, we can explicitly leverage evidential learning for uncertainty estimation in DRL action choice. The prediction $\hat{y}$ upon observation $x$ is equal to the expectation of the mean of the learned evidential distribution (i.e., $\gamma$). Aleatoric uncertainty can be computed as the expectation of the evidential variance while epistemic uncertainty can be computed as the variance in the mean which is intuitively the expected evidential variance divided by the \emph{virtual observations} $v$:

\begin{align}
\underbrace{\mathbb{E}[\mu]=\gamma}_{\text{prediction $\hat{y}$}};~\underbrace{\mathbb{E}[\sigma^{2}]=\tfrac{\beta}{\alpha-1}}_{\text{aleatoric uncertainty}};~\underbrace{\text{Var}[\mu]=\tfrac{\beta}{v(\alpha-1)}}_{\text{epistemic
uncertainty}}\text{.}
\end{align}

\begin{figure}[!t]
  \centering
  \includegraphics[width=\linewidth]{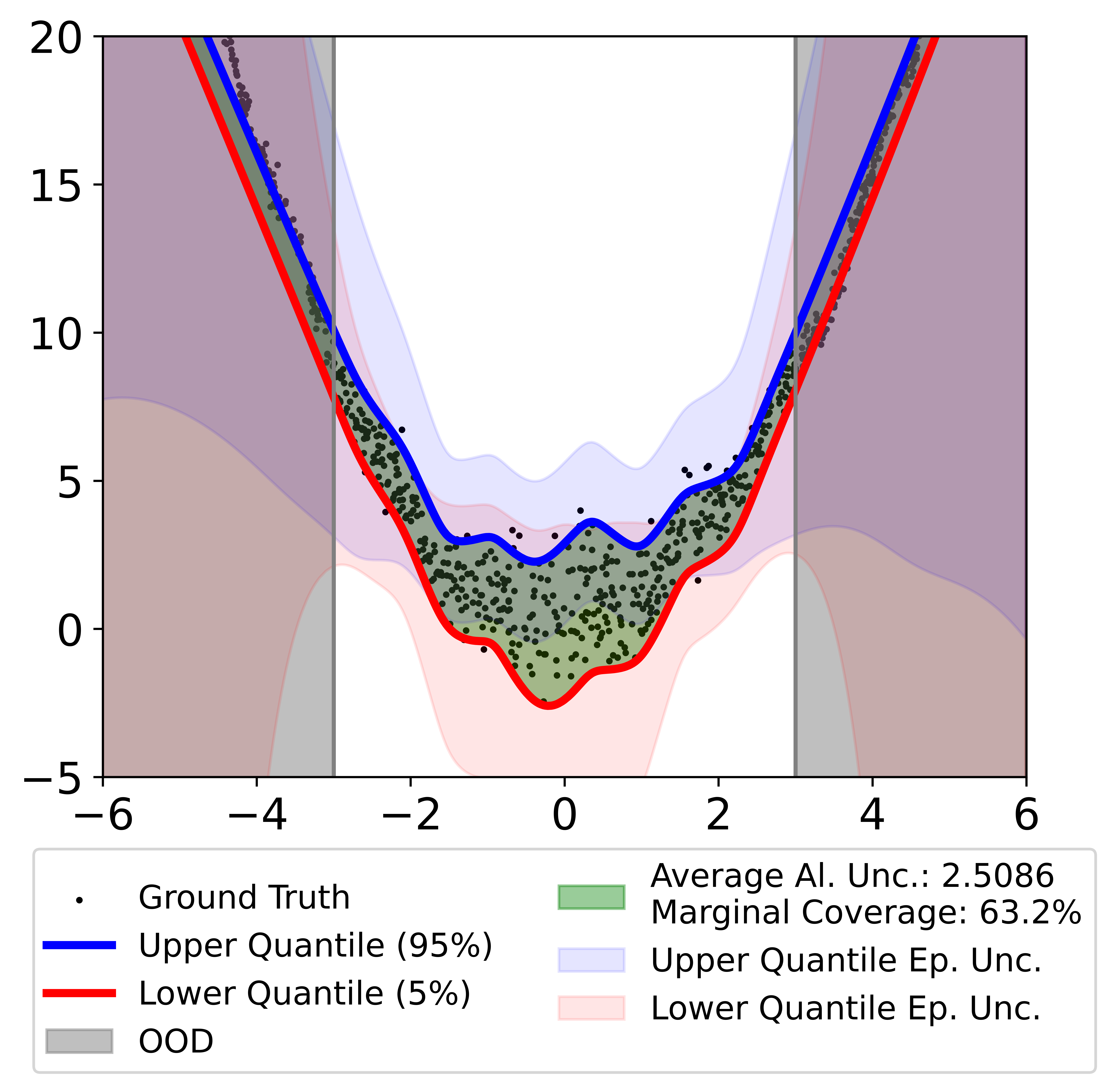}
  \caption{\textbf{Calibrated Evidential Quantile Regression Synthetic Example:} Demonstration of aleatoric and epistemic uncertainty estimates by the proposed algorithm on a fresh test set including OOD samples from function \mbox{$f(x) = \sin(3x) \cdot \cos(2x) + 0.5 \cdot e^{-x^2} + x^2 - 0.1x$} with added random noise $n\sim\mathcal{N}(0,1.5e^{-0.4|x|})$.}
  \label{uncertainty_example}
\end{figure}

A visual example of these quantities estimated within a relatively complex synthetic dataset (consisting of $1000$ unseen samples) using calibrated quantile regression is provided in Fig.~\ref{uncertainty_example}. It is shown that while the predicted quantiles are well calibrated around the noisy data, the underlying uncertainty associated with estimating them is appropriately moderate in the range $[-3,3]$ where the model was trained, and highly inflated for OOD samples outside this range. Accordingly, the presence of OOD samples hinders the model's ability to achieve the desired marginal coverage of $90\%$. This example represents the complementary relationship of calibrated quantile regression and evidential learning in accurately covering a complicated distribution of targets while clearly indicating a lack of confidence, especially in samples the model had not seen before.  In Section~\ref{sec:calibration}, we discuss the methods for the calibration and how it is applied to the predicted target distribution and evidential quantiles in DRL.

\subsection{Quantile Calibration via Conformal Methods}\label{sec:calibration}
Quantile regression is an inherently uncertainty-aware algorithm, which makes it attractive for integrating uncertainty into distributional reinforcement learning frameworks. However, quantile regression alone, while highly adaptive to heteroskedastic data, provides no guarantee of reasonable marginal coverage of a target $y$ from the target posterior distribution $Y$. Therefore, it is desired that the network estimating the conditional quantiles incorporate some method for quantile calibration as an added layer of protection against inaccurate predictions. This is particularly a challenge in reinforcement learning since samples may not necessarily be i.i.d., as is the primary assumption in standard supervised learning. In fact, to begin with, the DQN algorithm is able to perform well because it uses experience replay \citep{lin1992} to randomly sample previous interactions and a separate network to approximate targets so as to reduce correlation between training samples and decouple the learning process \citep{Mnih2015}. Therefore, any quantile calibration technique utilized must be performed as another training objective that does not degrade the integrity of the reinforcement learning process.

To this end, we adapt a conformal inference-based calibration method proposed in \citep{stutts2023} and referred to as conformalized joint prediction (CJP). Given inputs $X$, this method jointly trains a network to simultaneously output multivariate point predictions $\hat{Y}$ and lower and upper bound conditional quantiles that form an uncertainty-aware confidence interval $C(X)$ based on a desired guaranteed $90\%$ marginal coverage of $Y$. This is accomplished with a calibration dataset, which in typical conformal inference would be split from the training set and leveraged afterward, but CJP uniquely asserts that the entire training dataset can be used for calibration when coverage is approximated and averaged over random training batches. However, the coverage guarantee may not hold up in reinforcement learning since there is an underlying assumption that $X$ and $Y$ belong to a sufficiently similar distribution. For instance, if the robot agent escaping the maze in Fig.~\ref{fig:overview} continued to explore paths that were perceivably foreign, the estimated marginal coverage of $Y$ by $C(X)$ would decrease substantially. Nonetheless, the calibration measures used to produce the uncertainty-based conditional quantiles can still be applied in RL during training as a guide post for observations that \emph{are} similar, without disrupting experience replay. This can be constructively paired with the evidential learning discussed in Section~\ref{sec:DEL}, fusing better quantile estimation with OOD protection. 

There are two sets of quantiles to be calibrated in the proposed algorithm: those approximating the target distribution $Z_{\theta}$, and those approximating the sample mean (i.e., $\gamma$) of the evidential distribution. Both procedures rely on adaptions of functions originally proposed in \citep{chung2021}. 

To calibrate the DQN target quantiles, we embed the following loss function into the training objective:
\begin{subequations}
\begin{equation}
\mathcal{L}_{cal} = (1-\lambda_{cal})\times\text{COV}_{obj}+\lambda_{cal}\times\text{SHARP}_{obj}\text{,}
\end{equation}
where
\begin{equation}
\begin{split}
\text{COV}_{obj} = \frac{1}{N}\sum_{i=1}^{N}~&\mathbb{I}\{p^{cov}_{avg} < q_{i}\}[(y_{i}-\hat{q_{i}})\mathbb{I}\{y_{i} > \hat{q_{i}}\}]+ \\ 
& \mathbb{I}\{p^{cov}_{avg} > q_{i}\}[(\hat{q_{i}}-y_{i})\mathbb{I}\{y_{i} < \hat{q_{i}}\}]~\text{,}
\end{split}
\end{equation}
\begin{equation}
\begin{split}
\text{SHARP}_{obj}= \frac{1}{N}\sum_{i=1}^{N}~&\mathbb{I}\{q_{i} \leq 0.5\}[(1-\hat{q}_{i}) - \hat{q}_{i}]+\\
& \mathbb{I}\{q_{i} > 0.5\}[\hat{q}_{i} - (1-\hat{q}_{i})]~\text{,}
\end{split}
\end{equation}
\end{subequations}
$q$ is the target quantile level, $\hat{q}$ is the predicted quantile, $p^{cov}_{avg}$ represents the average probability that $y$ falls under $\hat{q}$, and $\lambda_{cal}$ is a hyperparameter balancing coverage of $y$ (COV$_{obj}$) with minimizing the distance between $\hat{q}$ and $1-\hat{q}$ in covering $y$ (SHARP$_{obj}$). This function individually calibrates each quantile predicted for $Z_{\theta}$ such that they are as close to the target levels as possible given success in training. 

For the evidential sample mean $\gamma$ we only predict and calibrate the $5^{th}$ and $95^{th}$ percentiles so as to establish a $90\%$ evidence-based confidence interval $C_{evi}(X)$ around the TD target from which we gauge uncertainty. However, because the computed TD target is of size $N$ quantiles, we must estimate $N$ quantiles of $\gamma$ for each percentile level. To establish the confidence interval as is done in \citep{stutts2023}, we reuse $\mathcal{L}_{cal}$ except that $q_{i}$ in the indicator functions is replaced with the desired marginal coverage rate $p=0.9$, ensuring that the focus of the calibration is on marginal coverage of the TD target over each training batch sampled from experience replay. Additionally, we seek to empirically center the TD target within the computed $C_{evi}(X)$ interval so as to enhance its accuracy. To do this, we add another loss function:
\begin{equation}
\begin{split}
\mathcal{L}_{interval} = \frac{1}{N}\sum_{i=1}^{N}~&(\hat{q}^{95^{th}}_{i}-~\hat{q}^{5^{th}}_{i})~+ \\
& \mathbb{I}\{y < \hat{q}^{5^{th}}_{i}\}\frac{2}{q}(\hat{q}^{5^{th}}_{i}-~y)~+ \\
& \mathbb{I}\{y > \hat{q}^{95^{th}}_{i}\}\frac{2}{q}(y-\hat{q}^{95^{th}}_{i})\text{.} \\
\end{split}
\end{equation}

These efforts to calibrate the quantiles related to $Z_{\theta}$ and evidential learning provide two major benefits. Calibrating the estimated quantiles of the value distribution allows the algorithm to achieve higher scores more quickly as a result of the refined approximation of $Z_{\theta}$. For evidential learning, the calibration results in more nuanced yet consistent uncertainty estimates that translate to better exploration strategies through action selection. A drawback associated with this is that individual tuning of hyperparameters is sometimes necessary for sufficiently distinct tasks in order to manage stability and maximize overall returns. Nevertheless, the advantages are worth the complexity overhead as shown in the results in Section~\ref{sec:results}.
\begin{figure*}[!t]
  \centering
  \includegraphics[width=\textwidth]{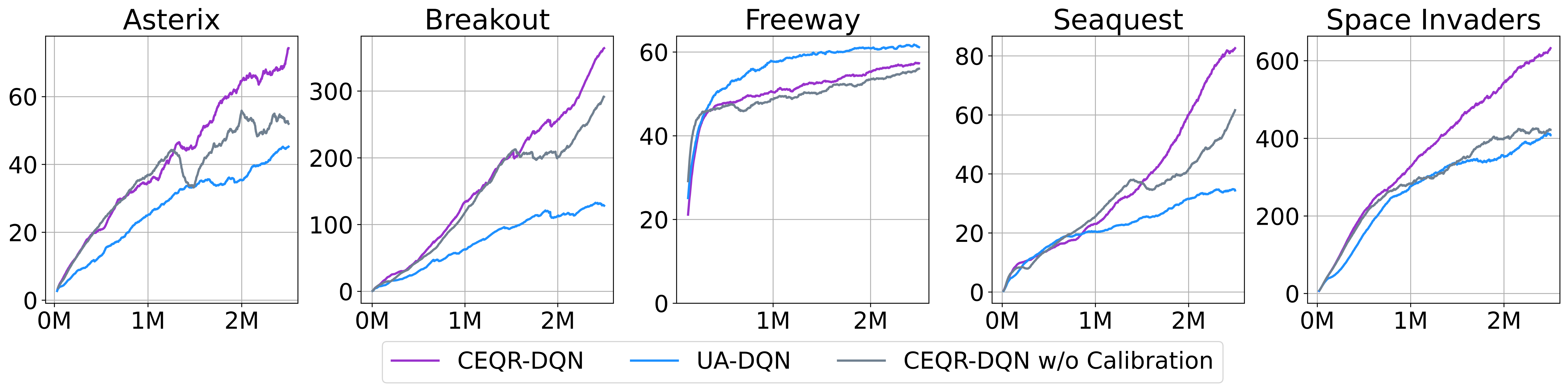}
  \caption{\textbf{MinAtar Results:} Comparison of second-order running average training scores between CEQR-DQN and UA-DQN across 25 random seeds on the simplified MinAtar games with $2.5$ million frames. Max episode scores for CEQR-DQN in each game are $158$, $4481$, $65$, $199$, and $4677$, while for UA-DQN they are $95$, $2457$, $69$, $95$, and $2231$. Results are also shown for CEQR-DQN without calibration.}
  \label{results}
\end{figure*}

\subsection{Calibrated Evidential Quantile Regression in DQN (CEQR-DQN)}\label{sec:compiled_theory}
To compile the theory in Sections~\ref{sec:DEL}~and~\ref{sec:calibration} into a brief practical tutorial for CEQR-DQN, we start with the model architecture shown in Fig.~\ref{architecture} in the appendix. In evaluating the algorithm on a miniaturized suite of Atari 2600 games (i.e., MinAtar \citep{young19minatar}) to directly compare with UA-DQN \citep{clements2020}, the trained network $\mathbb{N}$ resembles a miniaturized version of the original DQN network proposed in \citep{Mnih2015}. A single-layer CNN feature extractor receives a $10\times10\times O$ state observation input $x\in X$, where $O$ is a varying number of channels amongst the games, and outputs 1600 features to be fed through separate network heads for actions $a\in A$ and the evidential parameters $G=\{\gamma,v,\alpha,\beta\}_{evi}$. Notably, the output for actions is of size $A\times N$ while the output for the evidential parameters is of size $4\times A\times 2\times N$, indicating that there are N estimated quantiles for the $5^{th}$ and $95^{th}$ percentiles of each evidential parameter for each action. Ultimately, a separate uncertainty estimate is computed for each individual action upon each observation. To ensure that $v$, $\beta$, and $\alpha+1$ are positive as required in Alg.~1, the \emph{softplus} function is used at the output. 

Following the rest of Alg.~1, the expectation of each action, as well as the standard deviation of the learned evidential distribution (i.e., $\sigma_{evi}$) for each action, is calculated upon an observation $x$. With $\sigma_{evi}$, we can calculate the epistemic uncertainty $u_{evi}$ around the $5^{th}$ and $95^{th}$ percentile of the evidential sample mean $\gamma_{evi}$ and take their average over all quantiles to be the total epistemic uncertainty associated with each action in approximating the target distribution $Z_{\theta}$. Similarly, total aleatoric uncertainty can be calculated as the average of the absolute difference between the $5^{th}$ and $95^{th}$ percentile of $\gamma_{evi}$ over all quantiles. 

We can then utilize Thompson sampling to produce an uncertainty-aware action set sample $S_{A}$ from a multivariate normal distribution of the action means $\mathcal{M}$ and covariance based on the total epistemic uncertainty $\Psi_{ep}$ associated with the actions multiplied by hyperparameter $\lambda_{ep}$ to manage its influence. Optionally, as demonstrated in \citep{clements2020}, we can subtract from the actions the total aleatoric uncertainty $\Psi_{al}$ multiplied by a similar hyperparameter $\lambda_{al}$ to encourage the agent to avoid environmental risks. Taking the \emph{argmax} of $S_{A}$ results in an action choice that is at that point believed to be the most efficient in maximizing the return. Intuitively, the amount of least uncertain actions taken in an episode is approximately equal to the amount of greedy actions within a $\pm~0.1\%$ difference.

The rest of the algorithm follows the original QR-DQN procedure as outlined earlier in this section. The total quantile regression loss to be minimized in relation to learning $Z_{\theta}$, including calibration, is
\begin{equation}
\mathcal{L}_{Z} = \mathcal{L}_{qr} + \mathcal{L}_{cal}\text{,}
\end{equation}
while the total calibrated evidential learning loss to be minimized is
\begin{equation}
\mathcal{L}_{EL} = \mathcal{L}_{evi} + \mathcal{L}_{cal} + \mathcal{L}_{interval}\text{.}
\end{equation}

\section{Results}\label{sec:results}
In this section we discuss results of CEQR-DQN on the miniaturized Atari testbed MinAtar \citep{young19minatar}, which provides simplified versions of the games \emph{Asterix}, \emph{Breakout}, \emph{Freeway}, \emph{Seaquest}, and \emph{Space Invaders} in comparison to the seminal Arcade Learning Environment \citep{bellemare2013}. The games are simplified by reducing their observation feature space by over $20\times$, enforcing a smaller action space, and modifying certain environmental parameters so as to encourage behavioral learning over representation learning. Effectively, it makes evaluating discrete DRL algorithms faster and more easily reproducible. Furthermore, it encourages increased stochasticity through randomized spawn points, difficulty ramping, sticky actions, \emph{etc.} However, we disable sticky actions, which forces the agent to repeat the previous action with $10\%$ probability, as they are counter-intuitive to uncertainty-aware exploration strategies and hinder our quantile calibration efforts due to such random actions being included in gradient descent. All other MinAtar settings were left default.

As is shown in Fig.~\ref{results}, our algorithm demonstrates considerable improvements in terms of learning speed and score over UA-DQN \citep{clements2020}, which set the previous high scores. The two algorithms share hyperparameters, which are provided in Table~\ref{tab:hyperparameters} in the appendix. These improvements indicate that statistically robust uncertainty quantification, including calibration and modulated confidence upon OOD observations, can significantly alter an agent's exploration strategy for the better despite the environment being partially observable as is the case in each game except for \emph{Breakout}. Furthermore, since each evidential distribution is learned by the network through consideration of all observations, their derived uncertainty quantities allow the agent to maintain \emph{global} uncertainty awareness in action choice as opposed to \emph{local} awareness based on variance between two continuously changing auxiliary networks. The empirical benefit of this is especially demonstrated in \emph{Seaquest}, where rewards are complicated to obtain. Since in both algorithms the uncertainty estimates factored into action selection are based on evaluating the distribution of returns $Z_{\theta}$, such a difficult reward structure decreases their efficacy, especially when localized to a single observation. On the other hand, in \emph{Freeway}, rewards are scarce, so the less complex a model is, the better the performance. Concurrently, disabling sticky actions deepens the reliance of each algorithm's performance on uncertainty awareness and further emphasizes the need for non-localized estimates. The agents in the algorithms each take between $5\%-30\%$ uncertain actions throughout training, yet the difference in results are appreciable.

\section{Conclusion}\label{sec:conclusion}
We presented a novel framework for quantifying, calibrating, and exploiting aleatoric and epistemic uncertainty in distributional reinforcement learning involving quantile regression-based deep Q networks. The proposed methodology, applied to a suite of simplified miniature Atari games, combines deep evidential learning with quantile calibration based on properties of conformal inference to significantly improve agent exploration through uncertainty-aware action selection. This approach encourages an agent to learn faster and seek greater rewards in each successive training episode. The underlying theory is both easily adaptable and scalable, enabling its application to any task or dataset that necessitates consideration of uncertainty, particularly in scenarios with frequent out-of-distribution (OOD) observations.
\begin{acknowledgements} 
    This work was supported in part by COGNISENSE, one of seven centers in JUMP 2.0, a Semiconductor Research Corporation (SRC) program sponsored by DARPA.
\end{acknowledgements}
\bibliography{references}
\newpage
\onecolumn
\title{Echoes of Socratic Doubt: Embracing Uncertainty in\\Calibrated Evidential Reinforcement Learning\\(Supplementary Material)}
\maketitle
\appendix
\section{Model Architecture}
\begin{figure}[!h]
  \centering
  \includegraphics[width=\linewidth]{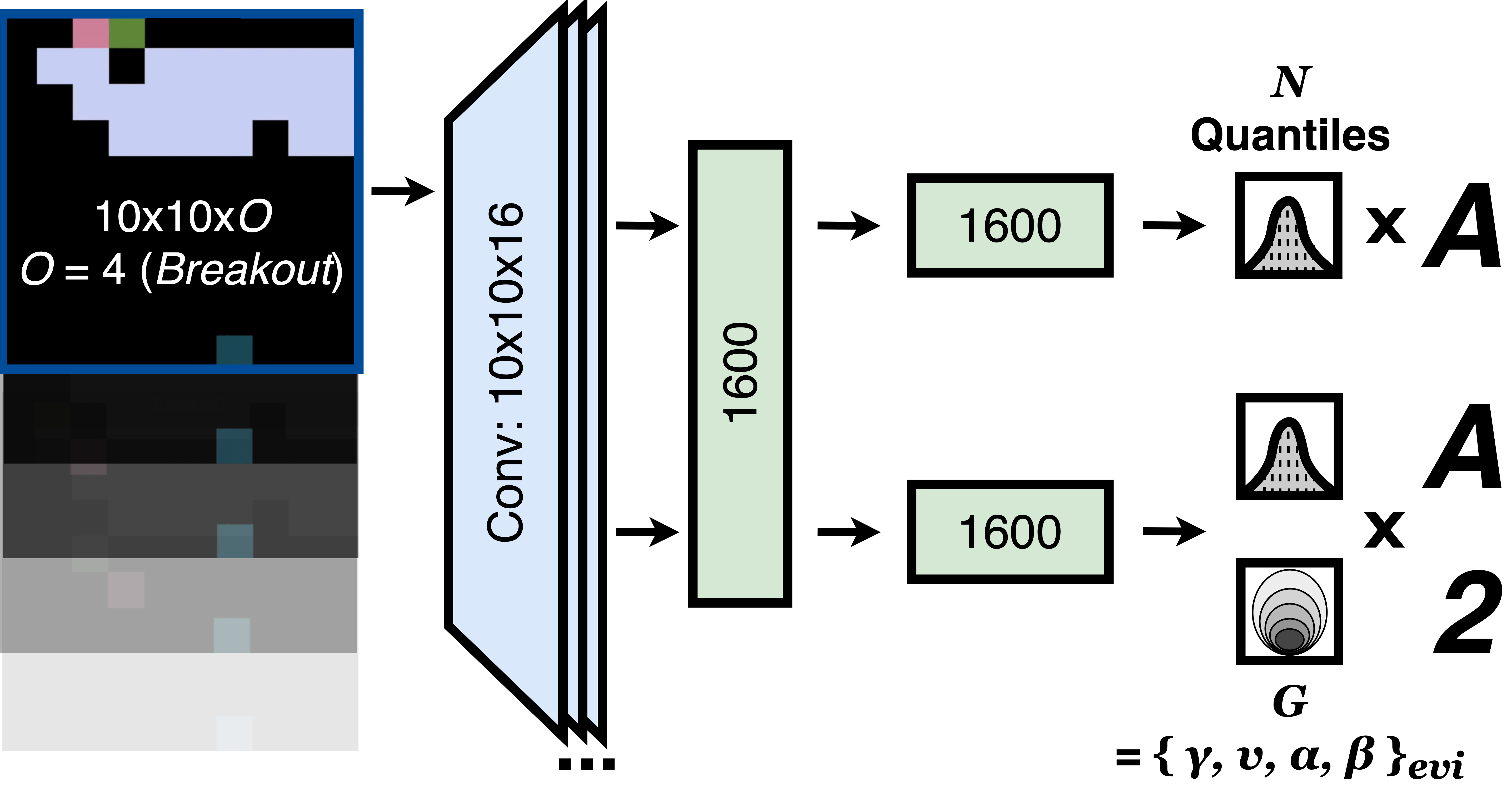}
  \caption{\textbf{Model Architecture for CEQR-DQN:} A single-layer CNN extracts features from an environment state observation and feeds two separate linear network heads to output $N$ quantiles for $A$ actions and the $5^{th}$ and $95^{th}$ percentiles for each evidential parameter associated with each action.}
  \label{architecture}
\end{figure}

Fig.~\ref{architecture} depicts the neural network model used for the proposed CEQR-DQN algorithm.

\section{Simulation Settings}
Table~\ref{tab:hyperparameters} lists the hyperparameters used by the proposed CEQR-DQN algorithm to achieve the results showcased in Section~\ref{sec:results} on MinAtar \citep{young19minatar}, a miniaturized suite of Atari 2600 games. We used code published by \citet{clements2020} to produce the results for UA-DQN. \textbf{Code for CEQR-DQN will be released at: \mbox{\url{https://github.com/acstutts/CEQR-DQN.git}}}

\begin{table}[!h]
    \centering
    \caption{Hyperparameters}\label{tab:hyperparameters}
    \begin{tabularx}{\columnwidth}{X|l}
        \toprule
        \bfseries Batch Size & 32\\
        \bfseries N Quantiles & 50\\
        \bfseries Replay Buffer Size & 100000 \\
        \bfseries Replay Start & 5000\\
        \bfseries Target~Network~Update~Frequency & 1000 \\
        \bfseries Discount Factor & 0.99\\
        \bfseries Training Frames & 2500000\\
        \bfseries Learning Rate & $10^{-4}$\\
        \bfseries Adam $\epsilon$ & $10^{-8}$ \\
        \bfseries Update Frequency & 1\\
        \bfseries $\kappa$ & 1\\
        \bfseries $\lambda_{al}$ & 0\\
        $\lambda_{el}~\{A,B,F,SQ,SI\}~(\text{CEQR-DQN})$ & $\{0.001,0.01,0.0002,0.005,0.0005\}$ \\
        $\lambda_{el}~(\text{UA-DQN})$ & $0.2$ \\
        \bfseries $\lambda_{reg}$ & 0.5 \\
        \bfseries $\lambda_{cal}$ & 0.5 \\
        \bottomrule
    \end{tabularx}
\end{table}
\end{document}